# A Domain-Adapted Pipeline for Structured Information Extraction from Police Incident Announcements on Social Media

Mengfan Shen[1], Kangqi Song[1], Xindi Wang[2], Wei Jia[3], Tao Wang[4], Ziqiang Han[1,5]

[1] School of Political Science and Public Administration, Shandong University, Qingdao, 266237, China

[2] School of Artificial Intelligence, Shandong University, Jinan, 250100, China

[3] School of Politics and Public Administration, Qingdao University, Qingdao, 266071, China

[4] School of International Affairs and Public Administration ,Ocean University of China, Qingdao, 266100, China

[5] Center for Crisis Research and Management, Tsinghua University, Beijing, 100084, China

Corresponding author: Ziqiang Han

Email: ziqiang.han@sdu.edu.cn

Address:Shandong University,QingDao,Huagang Buliding

## Abstract

Structured information extraction from police incident announcements is crucial for timely, accurate data processing, yet it poses considerable challenges due to the variability and informal nature of textual sources such as social media posts. To address these challenges, we developed a domain-adapted extraction pipeline that leverages targeted prompt engineering with parameter-efficient fine-tuning of the Qwen2.5-7B model using Low-Rank Adaptation (LoRA). This approach enables the model to handle noisy, heterogeneous text while reliably extracting 15 key fields, including location, event characteristics, and impact assessment, from a high-quality, manually annotated dataset of 4,933 instances derived from 27,822 police briefing posts on Chinese Weibo (2019-2020). Experimental results demonstrated that LoRA-based fine-tuning significantly improved performance over both the base and instruction-tuned models, achieving an accuracy exceeding 98.36% for mortality detection and Exact Match Rates of 95.31% for fatality counts and 95.54% for province-level location extraction. The proposed pipeline thus provides a validated and efficient solution for multi-task structured information extraction in specialized domains, offering a practical framework for transforming unstructured text into reliable structured data in social science research.

**Keywords:** LLM, Fine-tuning, Structured Information Extraction, Police Incident Announcements, NLP, Lora

## 1. Introduction

Crime data serve as a cornerstone of criminological research and evidence-based policy-making, but there is a lack of large-scale crime data from China. Researchers and practitioners can identify the spatial and temporal patterns of crimes, test theoretical frameworks such as routine activity theory, and evaluate the effectiveness of interventions by analyzing crime statistics (Eck & Weisburd, 2015). For instance, hotspot mapping of crime data has enabled police departments to adopt focused deterrence strategies, significantly reducing violent crime in urban areas (Braga et al., 2019). Longitudinal datasets, such as the National Crime Victimization Survey (NCVS) in the United States, reveal disparities between reported and unreported crimes, thereby refining the understanding of the "dark figures." (Roberts jr, 2010). Moreover, crime data underpins predictive policing models, though ethical concerns about algorithmic bias persist (Ferguson, 2017). Cross-national databases such as Eurostat and UNODC (United Nations Office on Drugs and Crime) surveys further facilitate comparative studies on the socioeconomic drivers of crime (Tseloni et al., 2010). Thus, robust crime data collection and transparent reporting are vital for advancing academic knowledge and shaping equitable safety policies.

However, large-scale, reliable, and publicly accessible crime datasets remain very limited in China, which hinders both the theoretical development of crime science and public safety practice. Only one recent study was identified (Zheng et al., 2024), which uses an AI-based method to extract community crime events with geographic coordinates and timestamps from judgment documents. More importantly, the judgment documents database (China Judgements Online) has recently slowed its update pace and cannot provide the most up-to-date information. It also does not involve non-criminal deviant behavior, as the judgment documents are from the court, while police briefings can be a valuable source for understanding the timely criminal events and patterns. The lack of large-scale crime data in China restricts empirical studies of crime trends, criminal behavior, and social impacts, leaving many vital questions underexplored in developing contexts (Yue et al., 2023). Therefore, there is a strong, urgent need to develop a structured crime-related dataset for social science research.

Police briefings, also known as crime incident announcements, are the primary official communication channels from police departments to the public in China and include basic information about criminal or deviant behavior incidents. These briefings, therefore, can be a critical but underdeveloped source of crime data for crime science research. As authoritative, timely public records, these police briefings can provide valuable data on spatiotemporal distributions of crime, incident characteristics, and severity assessments. However, because police briefings are presented in unstructured narrative form, they are not readily machine-readable and remain difficult to process (Fu et al., 2025). Unstructured text makes it challenging to extract consistent quantitative variables such as time, location, and incident outcomes, thereby limiting their analytical utility (Spicer et al., 2016).

Early and widely used methods for extracting crime-related information from social media have evolved through several distinct paradigms. Initially, research relied heavily on Feature Engineering and Classical Machine Learning. Techniques such as Support Vector Machines (SVM), Decision Trees, Random Forests, and Naïve Bayes were commonly applied to classify and identify malicious content or criminal behavior (Shafi et al., 2021). The performance of these models was contingent on extensive feature extraction, often utilizing methods like Term Frequency-Inverse Document Frequency (TF-IDF) for text representation and statistical methods such as the Gini Index or Chi-square for feature selection (Prathap et al., 2021), the Principal Component Analysis (PCA) for dimensionality reduction and computational efficiency improvement (Aghababaei & Makrehchi, 2016; Patel et al., 2025). To address the challenges of noisy and limited social media data, Rule-Based and Hybrid Systems were developed. These approaches integrated the aforementioned machine learning classifiers with expert-crafted logical rules, thereby improving precision in extracting specific emergency- or crime-related information (Shen et al., 2023). Subsequently, non-LLM Deep Learning architectures, including Convolutional Neural Networks (CNNs) and Recurrent Neural Networks (RNNs), offered significant advances by automatically learning feature representations (Tam & Tanrıöver, 2023). These models proved more effective for complex tasks such as sentiment analysis, topic modeling, and entity recognition, thereby enhancing the detection of nuanced criminal activity and behavioral patterns (Devarajan et al., 2024). Building upon the capabilities of deep learning, Data Fusion and Multimodal Analysis frameworks emerged. These methods combined social media data with external sources such as police records and Geographic Information System (GIS) data, significantly improving the accuracy of crime prediction and hotspot identification (Yang et al., 2017).

The conventional machine learning and deep learning methods are inadequate to address our tasks of turning the unstructured police briefings into structured information, because they often fail to enforce structured outputs, coordinate multi-task learning, or adapt to specialized domains (Raffel et al., 2020). Extraction information from unstructured police announcements and transform them into structured representations by identifying entities, relations, and events, is particularly demanding: models must produce strictly formatted outputs (Ching et al., 2018), handle multiple interdependent fields simultaneously, and capture domain-specific legal and criminological nuances.

Recent advances in large language models (LLMs) offer a path to overcoming these challenges. LLMs built on the Transformer architecture (Vaswani et al., 2017) exhibit strong semantic comprehension and text-generation capabilities, enabling them to process noisy, variable inputs. Models such as GPT (Radford et al., 2018) and BERT (Devlin et al., 2019) demonstrate strong semantic comprehension and generation abilities, making them particularly well-suited to processing variables in ambiguous text, such as the police briefings. These capabilities position LLMs as promising tools for overcoming the challenges of structured information extraction, including producing strictly formatted outputs, coordinating multiple subtasks, and interpreting domain-specific terminology.

Beyond their general capabilities, LLMs have introduced practical strategies for task adaptation, most notably prompt engineering and parameter-efficient fine-tuning. These approaches have become widely adopted because they provide flexibility and efficiency in adapting pre-trained models to downstream applications (Xu et al., 2023; Dagdelen et al., 2024; Xu et al., 2025; Chen et al., 2025). Prior research further underscores their potential for domain-specific information extraction: joint learning frameworks have been proposed for entity and relation extraction in classical Chinese texts (Tang et al., 2026), synthetic training data has been employed to address sparsity in triple extraction tasks (Guo et al., 2025), and few-shot prompting has yielded effective results in technical fields such as engineering (Aggarwal et al., 2026) and public policy analysis (Anglin et al., 2025). These studies demonstrate the versatility of LLMs but also highlight that specialized techniques remain necessary for achieving high accuracy in resource-constrained, domain-specific contexts.

Therefore, we integrate task-specific prompt engineering with Low-Rank Adaptation (LoRA) fine-tuning to meet the requirements of structured information extraction from police briefings. Prompt engineering enforces strict output formatting and guides the model to handle multiple fields consistently without modifying parameters (Lester et al., 2021; Liu et al., 2023). LoRA, a parameter-efficient fine-tuning method, introduces a small set of trainable parameters through low-rank matrix decomposition, enabling effective domain adaptation with minimal computational overhead (Howard & Ruder, 2018; Hu et al., 2022). Compared with alternatives such as adapters and prefix-tuning (Houlsby et al., 2019), LoRA offers a strong balance between efficiency, scalability, and empirical performance for our task. Together, these techniques form the technical foundation of our proposed pipeline for transforming unstructured police briefings into structured, analyzable data.

Establishing efficient information extraction mechanisms to transform textual police briefings into structured, analyzable data would substantially benefit both academic research and public safety practice. In particular, automated extraction of temporal and spatial references, event characteristics, and consequences is needed to unlock the analytical potential of these texts. Therefore, this study addresses the methodological challenge of converting unstructured police briefings into structured data suitable for computational analysis. Specifically, the objectives of this research are to:

- Develop a domain-adapted pipeline that transforms unstructured police briefings into structured datasets.
- Construct a high-quality, manually annotated dataset of Chinese police briefings posts to support model training and evaluation.
- Integrate task-specific prompt engineering with Low-Rank Adaptation (LoRA) fine-tuning to improve extraction accuracy and efficiency.
- Benchmark the proposed approach against baseline and instruction-tuned models to assess performance, consistency, and cost-effectiveness rigorously.
- Provide a practical and scalable methodology that enables researchers with limited technical resources to leverage social media–based police briefings for criminological and policy research.

This paper contributes to social computational research both methodologically and theoretically in the following three aspects:

First, we develop a cost-effective, domain-adapted information extraction pipeline that integrates task-specific prompt engineering with LoRA fine-tuning. Built upon the Qwen2.5-7B model, our approach effectively addresses the ambiguity, narrative variability and informal linguistic patterns in social media texts. By enforcing structured outputs while substantially reducing computational overhead, the pipeline lowers technical and financial barriers for researchers working under resource constraints.

Second, we construct a high-quality, manually annotated dataset comprising 4,933 instances sampled from 27,822 police briefings posted on Chinese Weibo between 2019 and 2020. This dataset directly mitigates the scarcity of reliable, domain-specific crime data in the Chinese context and provides a rigorously curated benchmark for structured information extraction in criminological and social media research.

Third, we conduct comprehensive empirical evaluations demonstrating the superiority of our pipeline over traditional baselines and instruction-tuned models. The system achieves 98.36% accuracy in mortality detection and exceeds 95% exact-match rates for both fatality counts and location extraction. These results validate a scalable framework for transforming unstructured social media narratives into structured, quantitative data suitable for computational social science analysis.

## 2. Method

In this study, we design a structured information extraction pipeline for police briefings, as illustrated in Figure 1. The pipeline consists of three main stages. In the data creation stage, we constructed a high-quality text dataset of police incident announcements by combining Python-based web crawling, OCR-based image-to-text conversion, regex-based normalization, duplicate removal, and length filtering, followed by double-verified manual annotation. During the training stage, we employed task-specific prompt engineering to construct dialogue-style training instances and fine-tuned the Qwen2.5-7B model (https://huggingface.co/Qwen/Qwen2.5-7B) with LoRA, thereby ensuring structured outputs and effective domain adaptation. Finally, in the evaluation stage, we compared the fine-tuned model against base and instruction-tuned baselines to assess its generation quality and task-specific performance across multiple extraction fields.

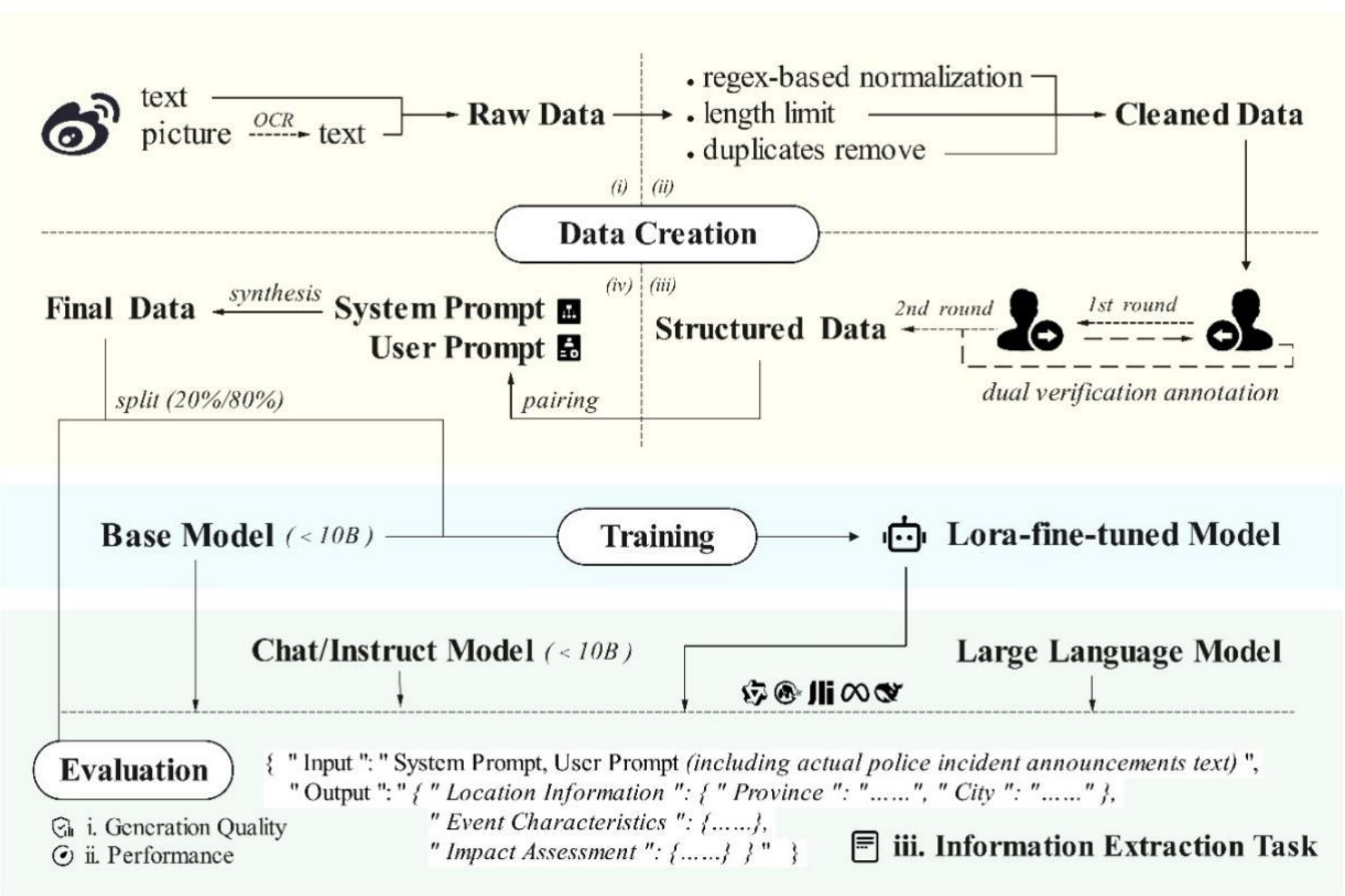

Fig.1 Overview of the proposed pipeline for structured information extraction from police announcements.

Note: The pipeline consists of three stages: (i) Data creation, including web crawling, OCR conversion, text cleaning, and dual-round manual annotation; (ii) Training, where structured prompts are paired with annotated data to fine-tune the Qwen2.5-7B model using Low-Rank Adaptation (LoRA); and (iii) Evaluation, where the fine-tuned model is benchmarked against base and instruction-tuned models on multiple extraction tasks, including location information, event characteristics, and impact assessment.

## 2.1 The Dataset Creation Stage

### 2.1.1 Data Collection

Since the “police announcement” or termed as “police situation announcement” or “police briefing” is the official public communications from police departments to the general public, we searched and identified all the Police Departments and the Political and Judiciary

Commission under the Committee of the Communist Party of China that have a verified Weibo account at all government levels, from the central government to the provincial, municipal/prefectural, and county/district government. According to the most recent information, there are 34 provinces (including Hong Kong, Macau, and Taiwan), 333 prefectures, and 2843 counties/districts in China; however, not all of these governments' police departments have verified Weibo accounts. After exhaustive searches and verification, we identified 3,969 official accounts associated with the Police Department or the Political and Judicial Commission at all government levels.

Since the "police briefing" is usually posted in text or image format, we developed a Python-based web crawler to collect the texts containing keywords such as "police briefings "police situation announcements," and "police announcements" posted by these accounts from January 1, 2019, to December 31, 2020, resulting in a total of 27,822 sample data. Each data record includes structured fields such as metadata (posting time, user ID), communication indicators (number of reposts, likes, comments), and content elements (text body and/or attached images). These accounts cover all 31 provincial-level administrative regions in China (excluding Hong Kong, Macao, and Taiwan). Textual information was extracted from the attached images using Optical Character Recognition (OCR) technology and saved in CSV files for subsequent processing.

#### 2.1.2 Data Cleaning and Preprocessing

The raw text underwent a rigorous cleaning pipeline: (1) Regex-based normalization preserving only Chinese characters, numerals, and standard punctuation while removing URLs and special symbols; (2) Manual screening revealed most content lacked case-specific information or contained insufficient data for structured extraction. Through empirical observation, we preliminarily filtered out texts containing fewer than 15 Chinese characters;

(3) Exact-match deduplication eliminated redundant entries. Manual review identified duplicated posts with inconsistent @User tags due to multi-account reposting, which were subsequently removed through pattern matching; (4) Since certain cases generated multiple follow-up reports over time, strict manual filtering was applied to ultimately yield 4,933 texts containing complete case information.

### 2.1.3 Data Annotation and Quality Assurance

Drawing on the codebook from previous studies on criminal behaviors from news reports or police announcements (Bowie, 2020; LaFree et al., 2022), the geospatial information, the case types, illegal behaviors, impacts of the behaviors, and police handling actions are selected as the primary topics (Uchida et al., 2024). In this analysis, we developed a codebook including 15 key variables for structured extraction, that encompasses location information (province, city), case type, illegal behaviors, police handling actions, the occurrence and specific numbers of casualties, the existence and precise amount of economic losses, involvement of cybercrime, completion status of the illegal act, case closure status, and assessment of social impact.

The annotation task was implemented through a rigorous manual process, involving three researchers with specialized backgrounds, to ensure the quality of the annotated gold-standard dataset. Specifically, two annotators were postgraduate students who had participated in multiple research projects in criminology and public safety management. Following the annotation guidelines detailed in Appendix 1, these two annotators independently completed the data labeling task using Excel spreadsheets. A dual-verification procedure was then conducted between the two annotators, and the Kappa consistency coefficient was calculated to be 94%—a result indicating excellent inter-annotator agreement.

The third participant, a professor and PhD in public crisis management and public policy, was responsible for the final double-check of contentious extraction items in which the two annotators had different rates. Such discrepancies were primarily concentrated in the textual description of "Illegal Behaviors and Police Handling Actions" as these categories involve relatively subjective judgments on language expression. After the professor's review and subsequent refinement of inconsistencies, the final high-accuracy, reliable annotated gold-standard dataset was formally generated.

### 2.1.4 Prompt Engineering and Training Data Synthesis

Prompt design significantly affects the effectiveness of subsequent fine-tuning processes, directly influencing the completion rate and accuracy of information extraction tasks. To achieve optimal model performance, a precise definition of extraction requirements and the standardization of the prompt output format are essential. In this study, we deployed the Qwen2.5-7b model locally using OLLaMA (Marcondes et al., 2025) for iterative prompt refinement and debugging. During this process, we revised the prompt content and structure by adjusting language, incorporating illustrative examples, and clarifying instructional details. Through multiple iterations of experimentation and evaluation, we developed a set of prompts characterized by comprehensive information coverage, clear structural organization, and explicit instructions. These robustly designed prompts provide a reliable foundation for the subsequent fine-tuning and structured information extraction tasks.

Table1. Prompts

| Prompt | Content |
| --- | --- |

<table>
<tr>
<td><b>System<br>Prompt</b></td>
<td>## Role Setting<br>You are a professional assistant for the structured extraction of information about policing situations. Please strictly extract information from the policing situation reports according to the following requirements.## Output Requirements<br>Please ensure that the output is in strict JSON format, including the following three parts:<br>1. Location information (province, city)<br>2. Event characteristics (type code, illegal means, etc.)<br>3. Impact assessment (casualties, losses, etc.)<br>### Event Type Coding Table<br>| Code | Type Description |<br>| ---- | ---- |<br>| 01 | Endangering national security |<br>| 02 | Endangering public safety |<br>| 03 | Economic and financial crimes |<br>| 04 | Infringement of personal rights |<br>| 05 | Infringement of property |<br>| 06 | Obstructing social management |<br>| 07 | Endangering national defense interests |<br>| 08 | Bribery and corruption |<br>| 09 | Dereliction of duty |<br>| 10 | Crimes committed by military personnel |<br>| 11 | Suicide |</td>
</tr>
</table>

| **User Prompt** | ### Please extract structured information from the following policing situation report:<br>×××**police incident announcements text**×××<br>### Data Extraction Requirements<br>1. **Location Information**:<br>-Province: Fill in the standard provincial name.<br>-City: Fill in the standard prefecture-level city name.<br>2. **Event Characteristics**:<br>-Case Type: Select from the following types (multiple selections are allowed).<br>-Illegal Means: Briefly describe.<br>-Cybercrime: true/false.<br>-Completed Illegal Act: true/false.<br>-Case Closure: true/false.<br>-Police Handling: Describe the handling measures.<br>3. **Impact Assessment**:<br>-Deaths: Existence (true/false) and the number of deaths.<br>-Injuries: Existence (true/false) and the number of injuries.<br>-Economic Losses: Existence (true/false) and the amount of loss (in yuan).<br>-Social Impact: true/false.<br>### Output Format Example<br>```json<br>{"Location":{"Province":"","City":""},"Event Characteristics":{"Type Code":[],"Illegal Means":"","Cybercrime":false,"Completed Illegal Act":false,"Case Closure":false,"Police Handling":""},"Impact Assessment":{"Deaths":{"Existence":false,"Number":0},"Injuries":{"Existence":false,"Number":0},"Economic Losses":{"Existence":false,"Amount":0},"Social Impact":false}} |
|---|---|

We conducted a dataset synthesis phase to integrate all processed data components after completing data collection, cleaning, annotation, and prompt engineering. Specifically, we merged the cleaned text data with the manually annotated structured information and incorporated the optimized prompts to ensure the dataset was well-suited for model training. As shown in Table 1, these are the final system prompts and user prompts we obtained after testing. The synthesized training data included the system prompt, the user prompt, the text content, and the assistant's output (in the form of manually structured JSON entries). The distinction between system prompts and user prompts is essential in LLM interaction design. System prompts, preset by designers, define the model’s identity, behavioral rules, and response boundaries, ensuring stable and coherent outputs. They serve as immutable global constraints that remain consistent across sessions. In contrast, user prompts are dynamic inputs that contain task-specific requests and require real-time model analysis. While system prompts establish the structural framework, user prompts serve as dynamic, task-specific directives within it. This process produced a unified, structured dataset in standard JSON format, providing a solid foundation for effective model fine-tuning and information extraction. Finally, we obtained 4,933 training dialogue samples, and the structure and format of the final training data are shown in Table 2.

Table2. JSON training format

| | |
|---|---|
| **Fine-tuning training set** | {"messages": [<br>{"role": "system",<br>"content:" **System Prompt** "},<br>{"role": "user", "content": " **User Prompt & police incident announcements text**"},<br>{ "role": "assistant","content": “**manually structured JSON entries** "}]} |

## 2.2 The Training Stage

The choice of a fine-tuning strategy requires careful consideration of computational constraints and trade-offs in model performance. Full-parameter fine-tuning, while theoretically optimal for task adaptation, proves prohibitively expensive for large language models, requiring O(n) memory for gradient computation where n exceeds billions of parameters. Adapter layers address this partially by introducing bottleneck architectures, but their sequential processing inherently increases latency during inference (Han et al., 2024). Prompt tuning eliminates parameter updates altogether, yet struggles with complex task specialization due to its limited representational capacity. Considering comprehensive factors, we chose Low-Rank Adaptation fine-tuning to ensure the quality of fine-tuning on the premise of low computational resource consumption (Lester et al., 2021).

### 2.2.1 Low-Rank Adaptation Fine-Tuning

LoRA is an efficient fine-tuning technique for pre-trained language models. It introduces a small number of trainable parameters via low-rank decomposition while keeping the original model parameters fixed. Specifically, for an initial pre-trained weight matrix $\mathbf{W_0}$ , LoRA approximates the weight updates $\mathbf{\Delta W}$ by factorizing them into two low-rank matrices **B and A**, where the Rank r of these matrices is significantly lower than the dimensions of $\mathbf{W_0}$, as illustrated in Figure 2. This design preserves the base model's capabilities while enabling targeted, task-specific adaptation.

In practice, LoRA employs a specialized initialization strategy: matrix **A** is initialized with random Gaussian values, while matrix **B** is initialized with zeros. This ensures that the model initially mirrors the pre-trained state, with incremental adaptation occurring during training. During forward propagation, the outputs from the original matrix ($\mathbf{W_0x}$) and the low-rank adaptation ($\mathbf{BAx}$) are combined to generate the final output. This approach maintains consistent input-output dimensions and enables efficient parameter updates via simple matrix operations.

LoRA offers several key advantages: (1) it significantly reduces the number of trainable parameters (typically less than 1% of the original model parameters), (2) it lowers computational overhead and memory usage, making it feasible to fine-tune large models on consumer-grade hardware, and (3) its modular design supports rapid switching between different task-specific adapters. Experiments demonstrate that LoRA effectively adapts models to domain-specific tasks while preserving their original performance.

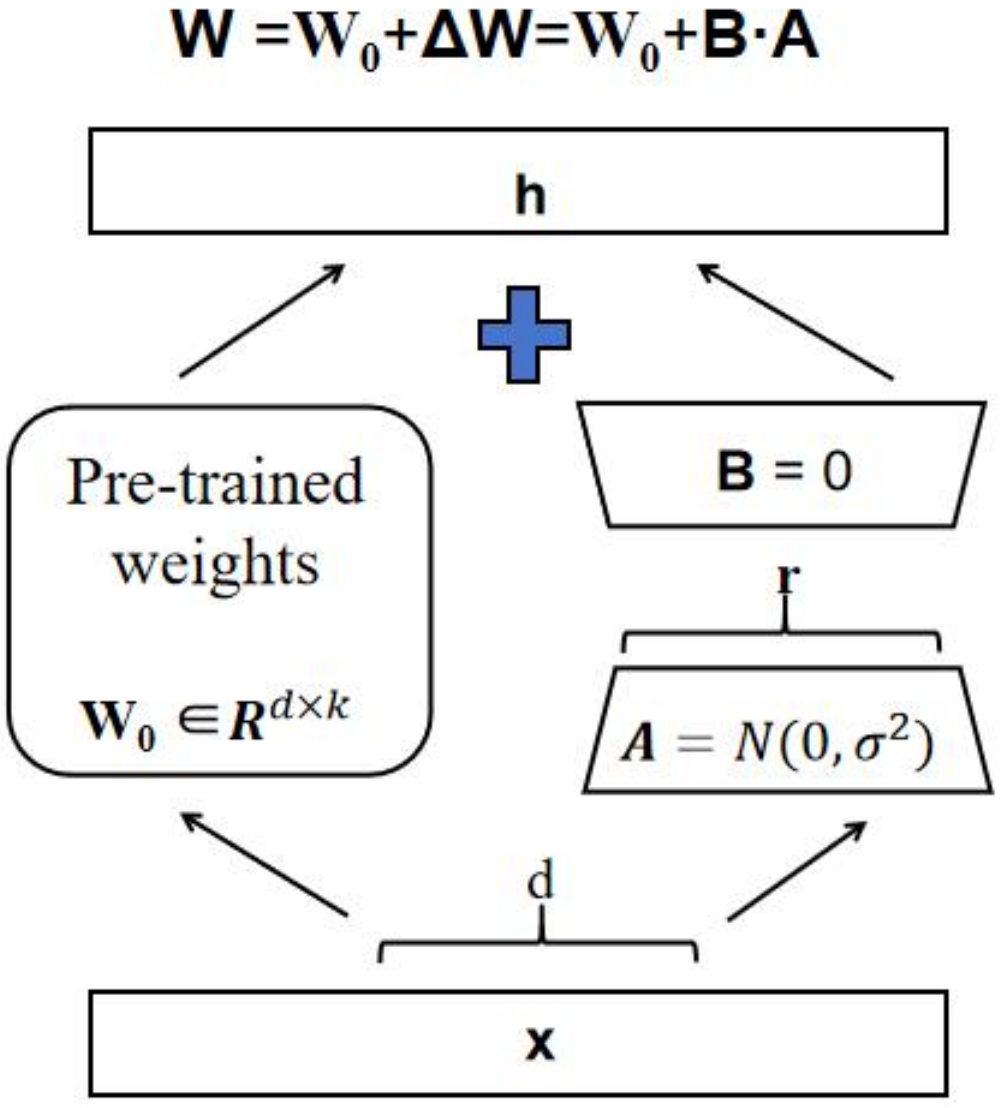

Fig.2 The training principle of LoRA fine-tuning

### 2.2.2 Fine-Tuning Tools and Model Selection

In this study, we selected Qwen2.5-7B-Base as the base model for LoRA fine-tuning, primarily because it is compatible with parameter-efficient fine-tuning methods and is licensed under the Apache 2.0 license, which facilitates both academic and commercial use. For comparative analysis, we included models released around the same period as Qwen2.5-7B, with similar parameter scales: BeiChuan2-7B, LLaMA3-CH-8B, Gemma2-9B, and ChatGLM2-6B, all of which were fine-tuned using LoRA. In addition, instruction-tuned versions (e.g., Instruct/Chat) of these models were included as baselines in subsequent experimental comparisons.

### 2.2.3 Fine-tuning Experimental Parameters

This study employs the LoRA (Low-Rank Adaptation) for parameter-efficient fine-tuning. Given the constrained sample size (fewer than 5,000 instances) and the fact that all text sequences were within 1,024 tokens, the following key training parameters were configured: the input sequence length was constrained to 1,024 tokens to maintain computational tractability. Training was performed for 8 epochs to ensure adequate model convergence, as indicated by the consistent descent and eventual stabilization of the training loss curve. As shown in Figure 3, larger learning rates triggered significant instability, whereas smaller ones led to relatively slow convergence. A rate of $1 \times 10^{-4}$ was found to be optimal, offering the best balance between stability and descent speed. This rate was therefore selected as the initial learning rate, alongside a dynamic learning rate scheduler. Besides, a per-GPU batch size of 4 was adopted with gradient accumulation over 8 steps, yielding an effective global batch size of 32.

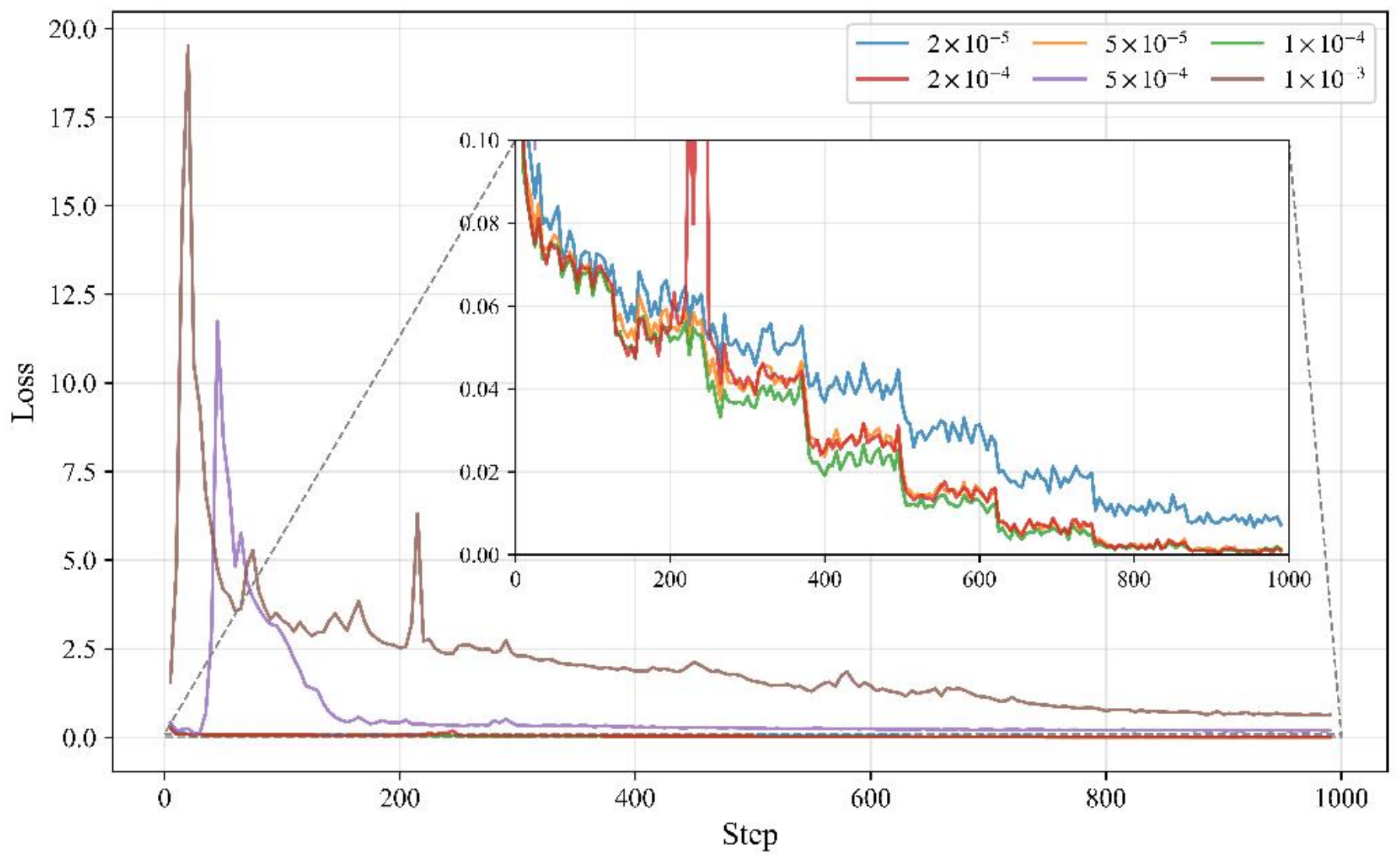

Fig.3 Training Loss Comparison Across Different Learning Rates.

Note: All other parameters are held constant as specified in the text to ensure a controlled comparison.

To obtain robust performance evaluation in the small-sample setting, a 5-fold cross-validation strategy was employed. Specifically, the dataset was randomly partitioned into five equal folds. In each iteration, four folds (80%) were used for training, and the remaining fold (20%) served as the test set. This process was repeated five times, with each fold used exactly once as the test set. The final model performance was reported as the average of the evaluation metrics obtained across all five test sets.

## 2.3 The Evaluation Stage

### 2.3.1 The Quality of Text Generation and Model Performance

To systematically evaluate the quality of text generation and model performance, we employed a suite of established metrics. Among these, BLEU-4 is a fundamental evaluation metric in natural language processing, measuring the similarity between generated and reference texts by computing the geometric mean of modified 4-gram precisions (Reiter, 2018). The formula for BLEU-4 is defined as:

$$BLEU-4 = \boldsymbol{BP} \times exp(\sum_{n=1}^{4} wn \cdot \quad log p_n) \quad (1)$$

where $p_n$ represents the precision of n-grams (n = 1,2,3,4), indicating the proportion of n-grams in the generated text that match the reference text; $w_n$ are weighting factors, typically set equally as $w_n = \frac{1}{4}$ to balance the influence of various n-gram lengths; and **BP** is the brevity penalty factor, which adjusts the score to prevent inflation caused by excessively short generated texts compared to the reference.

ROUGE-1, ROUGE-2, and ROUGE-L are Recall-Oriented Understudy for Gisting Evaluation (ROUGE) metrics (Lin, 2004)—a family of recall-focused metrics designed to assess the overlap between generated and reference texts. ROUGE-2 focuses on bigram recall, evaluating the co-occurrence of consecutive word pairs and thereby capturing semantic and syntactic similarity. ROUGE-L is based on the Longest Common Subsequence (LCS) between the generated and reference texts. It evaluates content similarity by considering the longest matching sequences, regardless of word order, to more comprehensively capture semantic coherence and overall textual similarity.

### 2.3.2 Evaluation Methods for Different Information Extraction Tasks

For structured information extraction tasks, we employed specialized evaluation metrics tailored to different data types. Boolean extraction was assessed using three complementary measures: Accuracy, which quantifies the overall correctness as the ratio of correct binary predictions to total cases; Recall, which evaluates the proportion of actual positive instances correctly detected by the model; and the F1-Score, providing a balanced assessment by calculating the harmonic mean of precision and recall, particularly valuable for datasets with imbalance.

Numerical data extraction performance is measured using the Exact Match Rate (EMR), which is the proportion of numerical values that exactly match the ground-truth references. This strict metric ensures absolute correctness, which is particularly critical in numerical fields where approximations are insufficient. Similarly, for location-related information such as province and city names, EMR is employed to evaluate precise matches of textual entries, as geographic entities require precise identification without semantic variations or synonyms. The consistent application of EMR across numerical and categorical location data provides a unified and rigorous assessment standard for critical fields in information extraction tasks. For descriptive text fields, we utilized Cosine Similarity calculated between TF-IDF vector representations of extracted and reference texts (Li & Han, 2013). This metric captures semantic similarity beyond surface-level string matching by measuring the angular proximity within vector space.

Categorical code extraction is evaluated using Jaccard Similarity (Equation 7), which compares the overlap between predicted and reference label sets (Bag et al., 2019). This set-based metric effectively handles unordered categorical assignments while accounting for partial matches.

$$Accuracy = \frac{TP+TN}{TP+FP+TN+FN} \tag{2}$$

$$Recall = \frac{TP}{TP+FN} \tag{3}$$

$$F1 = \frac{2 \times TP}{2 \times TP+FP+FN} \tag{4}$$

where:

TP = True Positives (correctly predicted positive cases)

TN = True Negatives (correctly predicted negative cases)

FP = False Positives (incorrectly predicted positive cases)

FN = False Negatives (incorrectly predicted negative cases)

$$EMR = \frac{Number\ of\ correctly\ predicted\ numerical\ values}{Total\ number\ of\ samples} \times 100\% \tag{5}$$

$$Similarity = \frac{\sum_{i=1}^{n} A_i B_i}{\sqrt{\sum_{i=1}^{n} {A_i}^2}\sqrt{\sum_{i=1}^{n} {B_i}^2}} \tag{6}$$

where: $Ai$ and $Bi$ are the *i-th* components of vectors ***A*** and ***B***, respectively, n is the dimensionality of the vector space

$$J(A,B) = \frac{|A \cap B|}{|A \cup B|} \tag{7}$$

where: A = Set of predicted labels (model output)

B = Set of ground truth labels (reference)

## 3. Results

A comparative analysis of fine-tuned models and their base and chat/instruction-tuned counterparts demonstrated substantial performance improvements from fine-tuning. To further assess the practical effectiveness of information extraction across different data types, we conducted indicator evaluations. The results indicated that fine-tuning provided minimal improvement for Boolean-type data, significantly improved text similarity scores for text data, and moderately improved performance on numerical data. Finally, benchmarking our fine-tuned Qwen-7B model against prominent LLMs, such as Qwen-Max, Gpt-5, and Deepseek-v3, revealed that a fine-tuned small-scale model typically achieved superior accuracy on our dataset. An example of the pipeline's extraction is shown in Figure 4.

**Input**:
**Prompt** +
"On October 24, a junior high school student in Renshou County, Meishan City, Sichuan Province, hit a teacher on the head with a brick in the classroom, causing severe injuries that required admission to the intensive care unit (ICU). On the 24th, a student named Yan (male, 15 years old) from a middle school in Renshou County attacked teacher Huang on the head with a brick in the classroom due to dissatisfaction with the teacher's daily management. Currently, Huang is receiving medical treatment, and Yan has been placed under criminal detention by the public security authorities. The case is under further investigation."

**Output**:
{"**Location**":{"**Province**":"Sichuan","**City**":"Meishan"},"**Event Characteristics**":{"**Type Code**":["04"],"**Illegal Means**":"A student struck a teacher's head with a brick","Cybercrime":false,"**Completed Illegal Act**":true,"**Case Closure**":false,"**Police Handling**":"The student has been placed under criminal detention by public security authorities"},"**Impact Assessment**":{"**Deaths**":{"Existence":false,"Number":0},"**Injuries**":{"Existence":true,"Number":1},"**Economic Losses**":{"Existence":false,"Amount":0},"**Social Impact**":False}}

**Input**:
**Prompt** +
"At approximately 9:25 a.m. on October 24, 2019, a traffic accident occurred at the intersection of Daduhe Road and Jinshajiang Road. Public security, emergency response, and 120 ambulance departments arrived quickly to handle the scene. Investigation revealed that the driver, Chen (male, 63 years old, a local resident), driving a Shanghai-licensed car, ran a red light at the intersection of Daduhe Road and Nujiang Road around 9:21 a.m., sideswiping a taxi and a woman riding an electric bicycle. He then accelerated to flee the scene. Upon reaching Jinshajiang Road, he again ran a red light through the intersection, colliding with an SUV that was making a normal right turn, and struck three non-motorized vehicles and several pedestrians. The driver, Chen, has been apprehended, and the case investigation is complete. Following the accident, leaders of the municipal party committee and government attached great importance, instructing relevant departments to make every effort to treat the injured. Currently, the accident has resulted in 5 deaths and 9 injuries. The injured are receiving full medical treatment.."

**Output**:
{"**Location**":{"**Province**":"Shanghai","**City**":"Shanghai"},"**Event Characteristics**":{"**Type Code**":["02"],"**Illegal Means**":"Driver ran red lights, fled after initial collision, and caused subsequent collisions resulting in casualties","**Cybercrime**":false,"**Completed Illegal Act**":true,"**Case Closure**":true,"**Police Handling**":"Driver apprehended. Case investigation is complete."},"**Impact Assessment**":{"**Deaths**":{"Existence":true,"Number":5},"**Injuries**":{"Existence":true,"Number":9},"**Economic Losses**":{"Existence":true,"Amount":"Not specified"},"**Social Impact**":true}}

Fig.4 Example cases illustrating the extraction pipeline

### 3.1 Comparative Performance of Lora-fine-tuned Models Across Different Model Variants

We evaluated four models: Qwen2.5-7B, BeiChuan2-7B, LLaMA3-CH-8B, and ChatGLM2-6B, in three configurations: base, LoRA-fine-tuned, and official instruct/chat-tuned. Text generation quality was measured with BLEU-4 and ROUGE-1/2/L, while computational performance was assessed with metrics such as throughput, step frequency, and initialization latency. Collectively, these metrics capture both the quality and efficiency of structured information extraction.

**LoRA vs. base models.** LoRA fine-tuning produced dramatic gains across all metrics, as shown in Figure 5. For instance, Qwen2.5-7B-LoRA achieved 93.76 in BLEU-4, compared with 24.97 for the base model, and 93.96 in ROUGE-1, compared with 40.05. BeiChuan2-7B improved from 14.09 to 93.72 (BLEU-4) and from 31.69 to 93.91 (ROUGE-1). Similar trends were observed for LLaMA3-CH-8B (from 6.99 to 91.36 in BLEU-4; from 28.23 to 91.61 in ROUGE-1) and ChatGLM2-6B (from 12.34 to 90.23 in BLEU-4; from 31.23 to 92.13 in ROUGE-1). Improvements were also observed for ROUGE-2 and ROUGE-L. Beyond scores, LoRA fine-tuning resolved major limitations of base models: while base versions often failed to comply with the required JSON schema, producing incomplete or inconsistent fields, LoRA-fine-tuned models generated outputs that were both schema-compliant and reliable for downstream analysis.

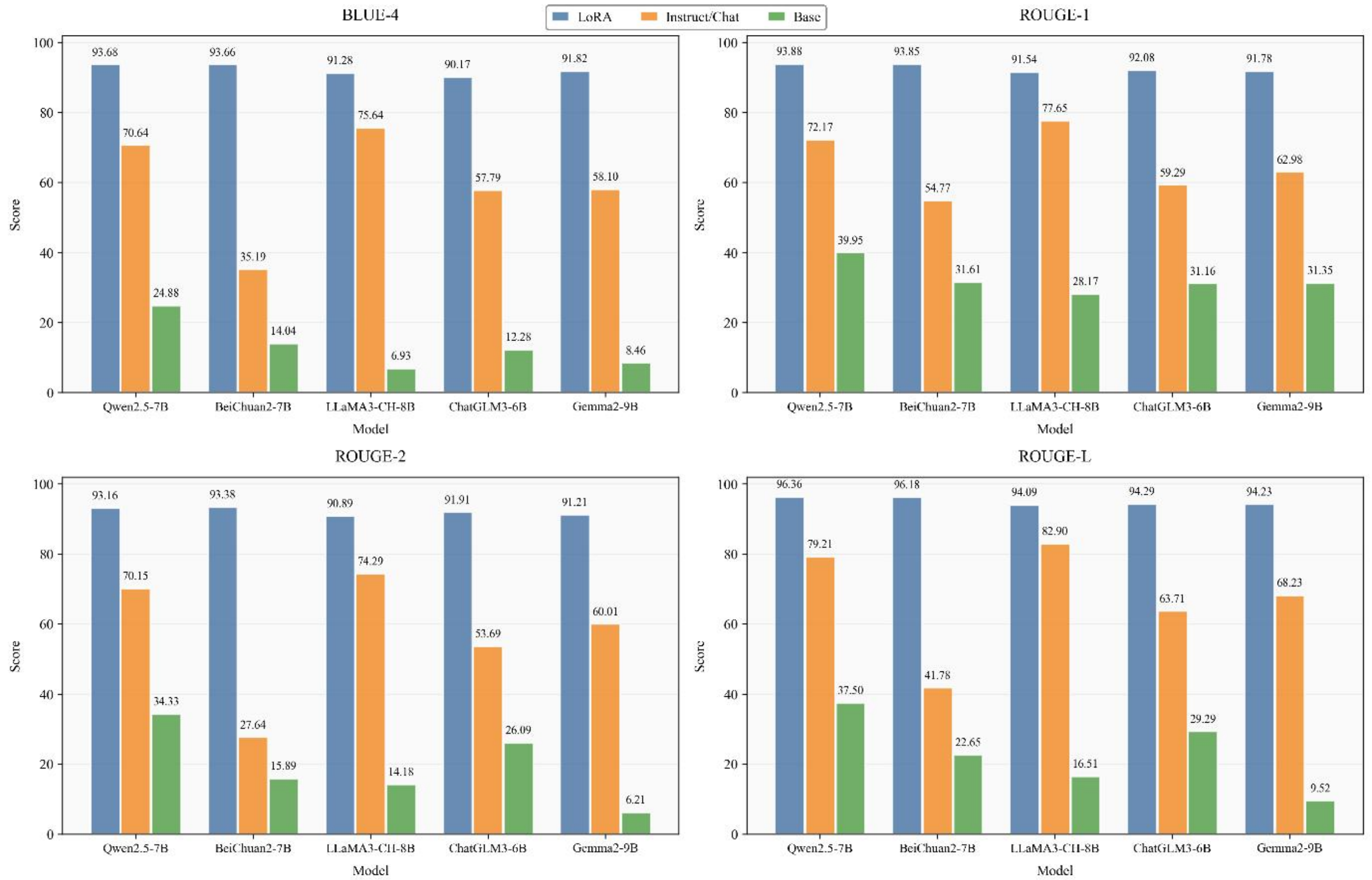

Fig. 5 Performance Comparison of Models (LoRA vs. Instruct/Chat vs. Base) evaluated by Bleu-4 and Rouge-L (1-2 & L) metrics. Additional detailed metrics are available in the Appendix

**LoRA vs. instruct/chat models.** LoRA-fine-tuned models also consistently outperformed official instruct/chat-tuned versions. For example, Qwen2.5-7B-LoRA achieved 93.76 in BLEU-4 versus 70.80 for Qwen2.5-7B-instruct, and 93.96 in ROUGE-1 versus 72.27. BeiChuan2-7B-LoRA surpassed its chat version, achieving 93.72 in BLEU-4 and 93.91 in ROUGE-1, compared to 35.29 and 54.86, respectively. LLaMA3-CH-8B and ChatGLM2-6B showed similar margins of improvement. Instruction-tuned models benefit from exposure to large-scale, generic corpora that enhance conversational ability, but their performance on structured tasks is limited by noisy training pairs and instruction mismatches. In contrast, our LoRA fine-tuning, grounded in a carefully curated, domain-specific, human-annotated dataset, yields far superior accuracy and consistency in extracting structured fields.

### 3.2 Performance Evaluation of Large Language Models in Information Extraction Tasks

While BLEU-4 and ROUGE provide useful indicators of overall text similarity, they are less effective for assessing structured information extraction, where precision on specific fields is critical. To ensure a comprehensive evaluation and validate the superiority of our fine-tuning approach against established benchmarks, we expanded our comparative analysis to include three traditional baselines: SVM+TF-IDF, TextCNN, and BERT-base-Chinese. Consequently, the evaluation encompasses three distinct model categories: traditional discriminative models, instruction-tuned LLMs (Qwen2.5-7B-instruct and LLaMA3-8B-CH-chat), and LoRA-tuned LLMs (including Qwen2.5-7B-LoRA, BeiChuan2-7B-LoRA, LLaMA3-8B-CH-LoRA, ChatGLM3-6B-LoRA, and Gemma2-9B-LoRA).

Boolean extraction. As shown in Table 3, the performance gap between traditional methods and fine-tuned LLMs is significant. Traditional models established a fundamental baseline but struggled with complex semantic judgments. SVM+TF-IDF showed the lowest performance, with accuracy hovering around 60-70% for most categories and dropping to 47.67% for social impact assessment. While BERT-base-Chinese improved upon these baselines—achieving 89.12% accuracy in mortality detection and 87.19% F1 score—it still lagged behind the generative models in capturing nuanced legal contexts.

Instruction-tuned models exhibited high variability across different fields. For instance, while Qwen2.5-7B-instruct achieved a competitive 91.72% accuracy in cybercrime detection, its performance plummeted in subjective categories, such as social impact (33.19%) and completed illegal acts (38.42%). Similarly, LLaMA3-8B-CH-chat showed a severe drop in the identification of economic losses (25.33%).

In contrast, LoRA-tuned models demonstrated consistently superior performance across all Boolean categories, significantly outperforming both traditional baselines and instruction-tuned counterparts. Qwen2.5-7B-LoRA reached 97.38% accuracy for mortality detection, 92.70% for injury identification, and 94.82% for completed illegal acts, with F1 scores of 96.82%, 90.91%, and 94.77%, respectively. Even in the challenging “Social impact” category, where traditional SVM (47.67%) and instruction-tuned models (<40%) failed, LoRA-tuned models maintained robust performance, with accuracy consistently exceeding 70% (e.g., BeiChuan2-7B-LoRA at 71.98%). Across all models, cybercrime detection consistently achieved strong results, yet the LoRA-tuned variants remained the most reliable, ranging from 92.84% to 94.63%.

Table 3. Performance Comparison of Models Across Multiple Tasks (Boolean Classification Tasks)

| Model | Evaluation | Death | Injury | Economic loss | Completed Illegal Act | Cybercrime | Social impact | Case closure |
|---|---|---|---|---|---|---|---|---|
| SVM+TF-IDF | Accuracy(%) | 69.45 | 71.23 | 66.54 | 72.15 | 74.32 | 47.67 | 74.89 |
| | Recall(%) | 65.12 | 65.45 | 60.12 | 65.34 | 70.15 | 43.23 | 68.45 |
| | F1(%) | 67.23 | 68.23 | 63.19 | 68.55 | 72.18 | 46.38 | 71.52 |
| TextCNN | Accuracy(%) | 79.34 | 75.67 | 72.45 | 66.78 | 78.90 | 62.34 | 74.56 |
| | Recall(%) | 75.12 | 68.23 | 65.12 | 58.45 | 72.34 | 45.67 | 68.12 |
| | F1(%) | 78.56 | 71.75 | 68.59 | 62.34 | 75.48 | 52.71 | 71.19 |
| BERT-base-Chinese | Accuracy(%) | 89.12 | 84.56 | 81.23 | 78.45 | 86.78 | 68.45 | 85.23 |
| | Recall(%) | 85.34 | 80.12 | 76.45 | 72.12 | 82.34 | 55.12 | 80.45 |
| | F1(%) | 87.19 | 82.28 | 78.76 | 75.15 | 84.50 | 61.07 | 82.77 |
| Qwen2.5-7B-instruct | Accuracy(%) | 83.12 | 72.35 | 87.3 | 38.42 | 91.72 | 33.19 | 71.74 |
| | Recall(%) | 77.24 | 66.73 | 81.95 | 33.56 | 86.92 | 28.40 | 66.50 |
| | F1(%) | 79.18 | 68.05 | 83.26 | 34.71 | 88.21 | 29.55 | 67.79 |
| LLaMA3-8B-CH-chat | Accuracy(%) | 77.93 | 72.83 | 25.33 | 55.21 | 81.85 | 37.99 | 67.27 |
| | Recall(%) | 72.88 | 68.42 | 20.08 | 49.87 | 76.58 | 32.68 | 61.4 |
| | F1(%) | 74.31 | 69.65 | 21.37 | 51.24 | 77.81 | 33.96 | 62.83 |
| ChatGLM3-6B-lora | Accuracy(%) | 96.84 | 90.52 | 88.45 | 93.51 | 92.91 | 69.34 | 93.21 |
| | Recall(%) | 92.69 | 86.88 | 83.15 | 89.76 | 88.36 | 58.94 | 89.48 |
| | F1(%) | 93.64 | 87.78 | 84.46 | 90.68 | 89.51 | 61.43 | 90.4 |
| LLaMA3-8B-CH-lora | Accuracy(%) | 97.23 | 92.67 | **89.76** | 94.63 | 93.55 | 70.55 | 94.27 |
| | Recall(%) | 96.85 | 91.82 | 88.23 | 93.32 | 90.46 | 65.78 | 93.85 |
| | F1(%) | 96.92 | 92.03 | 88.61 | 93.64 | 91.22 | 66.96 | 93.95 |
| Gemma2-9B-lora | Accuracy(%) | 97.26 | 90.51 | 86.63 | 92.6 | 92.84 | **71.46** | 89.34 |
| | Recall(%) | 96.31 | 88.26 | 83.77 | 90.16 | 90.53 | 78.42 | 90.54 |
| | F1(%) | 96.72 | 89.37 | 85.17 | 91.36 | 91.67 | 76.54 | 89.62 |
| BeiChuan2-7B-lora | Accuracy(%) | 97.19 | 92.41 | 89.68 | 94.11 | 93.27 | 71.98 | 92.94 |
| | Recall(%) | 96.54 | 89.9 | 87.36 | 92.91 | 91.88 | 68.53 | 90.52 |
| | F1(%) | 96.7 | 90.53 | 87.94 | 93.21 | 92.23 | 69.38 | 91.12 |
| **Qwen2.5-7B-lora** | Accuracy(%) | **97.38** | **92.70** | 89.75 | **94.82** | **93.68** | 71.15 | **94.39** |
| | Recall(%) | 96.63 | 90.31 | 86.63 | 93.40 | 92.33 | 67.35 | 92.82 |
| | F1(%) | 96.82 | 90.91 | 87.40 | 94.77 | 92.66 | 68.18 | 93.21 |

Numerical extraction. As illustrated in Table 4, to benchmark quantitative field extraction, we used BiLSTM-CRF and BERT-BiLSTM-CRF as traditional baselines. These sequence labeling models exhibited limitations in capturing numerical entities embedded within complex legal narratives. BiLSTM-CRF performed poorly, with Exact Match Rates (EMR) as low as 45.24% for fatality counts and just 25.12% for economic losses. While the introduction of pre-trained embeddings in BERT-BiLSTM-CRF improved performance, raising mortality extraction to 68.45%, it remained significantly less accurate than generative approaches.

In sharp contrast, LoRA-tuned models demonstrated superior capability in numerical precision. Qwen2.5-7B-LoRA achieved EMRs of 96.28% for fatality counts, 93.62% for injury counts, and 88.69% for economic loss. BeiChuan2-7B-LoRA followed closely, achieving 94.58%, 93.67%, and 87.15%, respectively. Other LoRA models maintained robust EMR scores, generally exceeding 80% across all metrics. Conversely, instruction-tuned models struggled significantly with specific numerical fields; notably, LLaMA3-8B-CH-chat recorded a substantial drop in performance, achieving only 11.52% accuracy in economic loss extraction, highlighting the need for parameter-efficient fine-tuning to achieve high precision for generating structured data.

Geographic extraction. The evaluation of geographic entity recognition, detailed in Table 4, further validates the advantages of the fine-tuned LLMs over traditional architectures. Traditional models struggled with granularity; BiLSTM-CRF achieved EMR scores of only 62.45% and 55.67% at the province and city levels, respectively. Although BERT-BiLSTM-CRF improved city-level recognition to 74.23%, it failed to match the contextual understanding of large models. Traditional extraction models may lack prior knowledge of China's administrative divisions and thus perform relatively poorly compared to LLMs.

LoRA-tuned models performed exceptionally well, particularly at the provincial level, with EMRs ranging from 89.92% (LLaMA3-8B-CH-LoRA) to 95.28% (BeiChuan2-7B-LoRA). Instruction-tuned models showed inconsistency, with Qwen2.5-7B-instruct performing well at the province level (86.31%) but dropping to 66.87% at the city level. At the more challenging city level, which requires finer discrimination, Qwen2.5-7B-LoRA achieved the highest EMR of 84.58%, significantly outperforming LLaMA3-8B-CH-chat, which reached only 40.41%. This demonstrates that while traditional models and standard instruction-tuning can handle broad geographic classification, LoRA-tuning provides the necessary refinement for precise multi-level location extraction.

Table 4. Performance Comparison of Models Across Multiple Tasks: EMR Evaluation Results

| Model | The number of deaths | The number of injured | The amount of the losses | Province | City |
|---|---|---|---|---|---|
| BiLSTM-CRF | 45.24 | 38.56 | 25.12 | 62.45 | 55.67 |
| BERT-BiLSTM-CRF | 68.45 | 62.12 | 68.34 | 68.56 | 74.23 |
| Qwen2.5-7B-instruct | 70.18 | 65.95 | 85.42 | 86.31 | 66.87 |
| LLaMA3-8B-CH-chat | 62.35 | 55.67 | 11.52 | 67.28 | 40.41 |
| ChatGLM-6B-lora | 92.16 | 88.42 | 86.28 | 91.15 | 80.65 |
| LLaMA3-8B-CH-lora | 91.28 | 86.73 | 83.15 | 89.92 | 72.84 |
| Gemma2-9B-lora | 93.52 | 91.45 | 81.36 | 92.18 | 67.42 |
| BeiChuan2-7B-lora | 94.58 | **93.67** | 87.15 | **95.28** | 82.69 |
| **Qwen2.5-7B-lora** | **96.28** | 93.62 | **88.69** | 94.43 | **84.58** |

Case type classification and textual extraction. As detailed in Table 5, for case-type classification, Qwen2.5-7B-LoRA again achieved the highest similarity score (82.55%), compared with 36.22% for Qwen2.5-7B-instruct. Performance across all models was more modest in this task, reflecting the subjective nature of case categories. For textual description extraction (police handling actions and criminal methods), Qwen2.5-7B-LoRA achieved similarity scores of 63.21% and 60.35%, whereas non-LoRA-tuned models scored below 20%. Although absolute scores appear low, they remain adequate for task completion since annotations were concise and textual variation often preserved semantic equivalence.

Table 5. Performance Comparison of Models Across Multiple Tasks (Case type and Text)

| Model | Case type | Police handling | Criminal methods |
|---|---|---|---|
| Evaluation | Jaccard Similarity | Cosine Similarity | Cosine Similarity |
| Qwen2.5-7B-instruct | 54.82 | 13.55 | 11.12 |
| LLaMA3-8B-CH-chat | 37.91 | 16.28 | 7.33 |
| ChatGLM-6B-lora | 79.68 | 57.14 | 54.27 |
| LLaMA3-8B-CH-lora | 81.25 | 55.89 | 52.76 |
| Gemma2-9B-lora | 80.45 | 63.08 | 61.22 |
| BeiChuan2-7B-lora | **83.42** | 62.07 | 61.78 |
| **Qwen2.5-7B-lora** | 81.33 | **65.51** | **63.14** |

### 3.3 Small Model Fine-tuning vs. Full-sized Large Models

To mitigate the deployment costs associated with large language models, two main approaches are often considered: fine-tuning models with fewer parameters and using API-based access to advanced models. This study compared the performance of fine-tuned smaller models with that of full-sized models to assess their relative effectiveness in structured information extraction.

We conducted experiments with two API-based models, Qwen-max, GPT-5, Claude-4, Genimi2.5-pro, and Deepseek v3, using consistent prompt engineering (matching the prompts and system instructions from previous experiments) and few-shot examples aligned with prior experiments. Input data were processed in streaming mode, and outputs were recorded for structured evaluation.

**Task-specific results.** Table 6 presents the performance across key fields, including Death, Injury, Economic loss, Crime success, Social impact, Cybercrime, and Case closure. Among the large-scale models, Gemini2.5-pro demonstrated superior performance on the Cybercrime and Social impact tasks compared with Qwen2.5-7B-LoRA. GPT-5 and Claude-4 also demonstrated competitive performance across several tasks, particularly in Crime success and Cybercrime. While Deepseek V3 and Qwen-max outperformed Qwen2.5-7B-LoRA on Economic loss and Social impact tasks, the fine-tuned Qwen2.5-7B-LoRA remained highly competitive across most evaluation metrics, achieving performance comparable to larger models on tasks such as Death and Case closure.

Table 6. Performance Comparison of **Small Model Fine-tuning vs. Full-sized Large Models** Across Multiple Tasks (1)

| **Model** | **Evaluation** | **Death** | **Injury** | **Economic loss** | **Crime success** | **Cybercrime** | **Social impact** | **Case closure** |
|---|---|---|---|---|---|---|---|---|
| | Accuracy(%) | 97.96 | 88.31 | **92.21** | 63.53 | 91.19 | 88.86 | 78.90 |
| Deepseek v3 | Recall(%) | 97.95 | 88.31 | 92.25 | 94.34 | 93.26 | 67.35 | 92.82 |
| | F1(%) | 98.96 | 93.79 | 93.74 | 95.73 | 93.6 | 68.18 | 93.21 |
| | Accuracy(%) | 98.25 | 92.85 | 89.45 | 94.62 | 94.32 | 78.35 | 93.28 |
| Gpt-5 | Recall(%) | 97.45 | 90.38 | 86.72 | 93.45 | 93.41 | 74.82 | 91.95 |
| | F1(%) | 97.62 | 91.18 | 87.85 | 94.86 | 94.78 | 75.24 | 92.38 |
| | Accuracy(%) | 98.18 | 92.47 | 89.83 | 94.25 | 93.42 | 79.64 | 92.91 |
| Claude-4 | Recall(%) | 97.32 | 89.95 | 86.28 | 92.87 | 91.23 | 76.18 | 91.42 |
| | F1(%) | 97.48 | 90.68 | 87.52 | 93.56 | 92.54 | 76.93 | 91.85 |
| | Accuracy(%) | 98.42 | 93.28 | 90.24 | 95.15 | **95.64** | 90.27 | 93.85 |
| Gemini2.5-pro | Recall(%) | 97.58 | 90.67 | 87.15 | 93.82 | 97.32 | 87.42 | 92.15 |
| | F1(%) | 97.75 | 91.45 | 88.12 | 94.48 | 97.76 | 88.15 | 92.68 |
| | Accuracy(%) | 94.15 | 84.86 | 88.96 | 77.44 | 93.90 | **91.60** | 78.25 |
| Qwen-max | Recall(%) | 94.08 | 84.32 | 87.67 | 77.38 | 89.72 | 87.43 | 79.1 |
| | F1(%) | 96.83 | 90.17 | 89.19 | 86.83 | 93.15 | 90.65 | 87.34 |
| | Accuracy(%) | **98.36** | **93.64** | 90.66 | **95.78** | 94.63 | 71.15 | **94.39** |
| **Qwen2.5-7B-lora** | Recall(%) | 97.61 | 91.22 | 87.51 | 94.34 | 93.26 | 67.35 | 92.82 |
| | F1(%) | 97.80 | 91.83 | 88.28 | 95.73 | 93.60 | 68.18 | 93.21 |

**Comparative analysis.** Tables 7 and 8 summarize results across multiple evaluation metrics for six models. The fine-tuned Qwen2.5-7B-LoRA demonstrated particularly strong performance in location extraction tasks (Province and City) and criminal analysis metrics (Police handling actions and Criminal methods), significantly outperforming all larger models by substantial margins. While Qwen-max and Deepseek v3 demonstrated strong general-purpose capabilities, and Gemini2.5-pro showed competitive results in basic extraction tasks (The number of deaths, injuries, and losses), the fine-tuned Qwen2.5-7B-LoRA remained highly competitive and, in most cases, outperformed the larger models on these domain-specific tasks. The substantial performance advantages in complex semantic understanding tasks suggest that targeted fine-tuning can enable smaller models to exceed the performance of much larger models in specialized extraction domains.

Table 7. Performance Comparison of Small Model Fine-tuning vs. Full-sized Large Models Across Multiple Tasks (2): EMR Evaluation Results

| Model | The number of deaths | The number of injured | The amount of the losses | Province | City |
|---|---|---|---|---|---|
| Deepseek v3 | 93.75 | 84.72 | 84.38 | 90.65 | 77.89 |
| Gpt-5 | 94.25 | 88.62 | 87.54 | 79/24 | 72/15 |
| Claude-4 | 93.2 | 86.45 | **91.43** | 76.34 | 73.23 |
| Gemini2.5-pro | **95.56** | 89.32 | **92.45** | 78.23 | 77.53 |
| Qwen-max | 92.71 | 80.87 | 88.38 | 79.44 | 81.81 |
| **Qwen2.5-7b-lora** | 95.31 | **92.55** | 89.78 | **95.54** | **85.73** |

Table 8. Performance Comparison of **Small Model Fine-tuning vs. Full-sized Large Models** Across Multiple Tasks (3)

| Model | Case type | Police handling | Criminal methods |
|---|---|---|---|
| **Evaluation** | **Jaccard Similarity** | **Cosine Similarity** | |
| Deepseek v3 | 78.16 | 17.45 | 8.25 |
| Gpt-5 | 69.43 | 21.23 | 16.34 |
| Claude-4 | 73.54 | 27.34 | 22.23 |
| Gemini2.5-pro | 77.53 | 33.23 | 26.87 |
| Qwen-max | 74.22 | 25.60 | 15.64 |
| **Qwen2.5-7b-lora** | **82.55** | **63.21** | **60.35** |

### 3.4 Ablation Study and Sensitivity Analysis

To rigorously validate the robustness of the proposed pipeline and dissect the contributions of its key components, we conducted a series of ablation studies and sensitivity analyses. Specifically, this section evaluates the impact of task-specific prompt engineering and assesses the model's stability under varying LoRA hyperparameters and decoding strategies.

#### 3.4.1 Ablation Study on Prompt Engineering

To verify the necessity of the iterative prompt refinement described in Section 2.1.4, we evaluated the impact of prompt design on extraction performance. Given that the base model lacks the instruction-following capabilities required for zero-shot extraction, we conducted this ablation study using the Qwen2.5-7B-Instruct model to ensure a fair comparison. We contrasted a Basic Prompt (generic natural language instructions without schema constraints) with our Structured Prompt (Table 1).

The quantitative results, summarized in Table 9, highlight a clear stepwise improvement. The introduction of the Structured Prompt alone yielded a marked gain over the Basic Prompt, raising the Average F1-Score from 58.32% to 67.25% and the Average EMR from 63.65% to 74.95%. This confirms that explicit constraints effectively guide the model's focus. However, the most significant leap was achieved by combining the structured prompt with LoRA fine-tuning, which outperformed the instruction-tuned baseline by a substantial margin (reaching 89.14% F1 and 91.52% in EMR). These findings demonstrate that while prompt engineering ensures structural compliance, fine-tuning is indispensable for high-precision domain adaptation.

Table 9. Ablation Study Results: Impact of Prompt Engineering and LoRA Fine-tuning on Classification and Extraction Performance

| Model | Prompt Strategy | Average F1-Score (%) | Location EMR (%) |
|---|---|---|---|
| Qwen2.5-7B-Instruct | Basic Prompt | 58.32 | 63.65 |
| Qwen2.5-7B-Instruct | Structured Prompt | 67.25 | 74.95 |
| **Qwen2.5-7b-lora** | Structured Prompt | **89.14** | **91.52** |

Note: "Average F1-Score" represents the mean F1 score across all Boolean classification tasks (e.g., Mortality, Cybercrime). "Average EMR" denotes the mean Exact Match Rate for all structured extraction fields, including numerical counts (deaths, injuries, losses) and geographic entities (province, city).

### 3.4.2 Sensitivity Analysis: Hyperparameters and Decoding Strategies

We conducted a rigorous sensitivity analysis on two critical dimensions: model capacity (determined by LoRA Rank r) and inference determinism (controlled by Temperature T).

Impact of LoRA Architecture (r). We evaluated ranks in {8, 16, 32, 64} with a fixed scaling factor alpha = 2r. As shown in the top section of Table 10, performance peaked at r=16, achieving an Average F1-Score of 89.14% and Average EMR of 91.52%. Lower ranks (r=8) resulted in underfitting (86.45% F1), while higher ranks (r=32, 64) yielded negligible gains or slight degradation, increasing computational cost without improving accuracy. Thus, r=16 was identified as the optimal configuration for parameter efficiency.

Impact of Decoding Temperature (T). We tested temperatures T in {0.1, 0.5, 0.9} to assess generation stability. The results reveal a strong negative correlation between temperature and performance. Low temperature (T=0.1) maximized precision (89.14% F1), whereas high temperature (T=0.9) caused a sharp decline to 79.25%, introducing hallucinations and syntax errors. Consequently, T=0.1 was adopted to ensure reproducibility and strict adherence to the schema.

Table 10. Sensitivity Analysis of LoRA Rank and Decoding Temperature on Extraction Performance

| **Parameter Type** | **Setting** | **Average F1-Score (%)** | **Average EMR (%)** |
|---|---|---|---|
| | r=8 (alpha=16) | 86.45 | 88.23 |
| **LoRA Rank(r)** | r=16 (alpha=32) | **89.14** | **91.52** |
| | r=32 (alpha=64) | 89.05 | 91.48 |
| | r=64 (alpha=128) | 88.92 | 91.10 |
| | T=0.1 | **89.14** | **91.52** |
| **Temperature(T)** | T=0.5 | 87.65 | 88.40 |
| | T=0.9 | 79.25 | 86.35 |

## 4. Discussion

### 4.1 Interpretation of Results

Across all experiments, LoRA fine-tuning substantially improved the performance of mid-sized models compared with their base versions. The fine-tuned models consistently produced outputs that were not only more accurate but also schema-compliant, overcoming the frequent formatting errors observed in base models. This highlights LoRA's effectiveness in enforcing structured output requirements, which are critical for downstream analysis of police briefings.

When compared with instruction- or chat-tuned models, LoRA-fine-tuned versions also showed clear advantages. Because they were trained on carefully curated, domain-specific annotations, they produced more reliable and consistent results than models trained on large but noisy general corpora. This suggests that quality-focused adaptation can outperform scale alone in domain-specific structured extraction tasks.

Furthermore, our comparative analysis reveals that fine-tuned LLMs significantly outperform traditional classification and extraction models, such as SVM, TextCNN, and BiLSTM-CRF. Unlike these classical approaches, which often rely on shallow feature engineering or limited contextual windows, LLMs leverage extensive prior knowledge and deep semantic understanding acquired during pre-training. This capability proves critical for information extraction tasks, enabling the model to resolve complex syntactic structures and infer implicit details, such as distinguishing specific economic losses from general numerical data or identifying administrative jurisdictions, where traditional sequence-labeling models frequently fail. This demonstrates that the generative paradigm, reinforced by domain adaptation, offers a superior mechanism for handling the variability and ambiguity inherent in unstructured police narratives.

Finally, compared with state-of-the-art large models accessible via APIs, LoRA-tuned smaller models achieved competitive performance across most objective extraction tasks, including event detection, outcome quantification, and location extraction. In tasks involving subjective judgment, such as social impact assessment, large language models (LLMs) outperform fine-tuned small models. When determining whether a case has caused a severe social impact, particularly when the text lacks explicit indications, the evaluation should comprehensively consider factors including the severity of the case, the extent of negative repercussions, the scale of affected populations, and the magnitude of incurred losses. By analyzing misjudged cases, we observe that fine-tuned small models exhibit limited capacity and are more susceptible to surface-level textual features, failing to conduct comprehensive multidimensional assessments.

### 4.2 Error Analysis

A qualitative analysis of the errors is reported in Table 11; more detailed cases are provided in the appendix.

Table 11. Summary of Model Error Analysis Across Different Extraction Tasks (Qwen2.5-7b-lora)

| Extraction Type | Specific Content | Error Analysis |
|---|---|---|
| **Location Information** | Province, City | Failed to identify the primary jurisdiction among multiple locations. |
| | | Misidentified administrative hierarchy, favoring sub-prefectural units. |
| | | Generated hallucinations when processing township-level information. |
| **Event Characteristics** | Case Type | Lacked legal expertise to differentiate crime types. |

| | | Over-generalized charges in legal classification. |
|---|---|---|
| | Completed Illegal Act | Misjudged the act of completion due to the absence of severe consequences. |
| | Cycercrime | Literally interpreted "network" as cybercrime. |
| | Case Closure | Misunderstood judicial procedures regarding case resolution. |
| | Illegal Means and Police Handling | Evaluation confirms that the model output is more faithful to the original and more detailed than manual annotations, without affecting subsequent analysis. |
| **Impact Assessment** | Deaths | Failed to distinguish case-related fatalities. |
| | | Miscounted deaths due to keyword recognition failures. |
| | Injuries | Overlooked implicit injury indicators in medical contexts. |
| | Economic Losses | Made calculation errors in multi-step problems. |
| | Social impact | Overestimated severity due to exaggerated descriptions. |

Note: Detailed case demonstrations are provided in the Appendix.

### 4.3 Theoretical Implications

This study contributes to the theoretical understanding of structured Information Extraction (IE) in three ways.

First, our results show that mid-sized LLMs with LoRA fine-tuning can match or exceed the performance of larger, fully fine-tuned or instruction-tuned models on structured extraction tasks. This challenges the assumption that only very large-scale models can deliver state-of-the-art accuracy for complex IE.

Second, the combination of prompt engineering with LoRA fine-tuning provides evidence for a hybrid adaptation strategy. While prompting guides the model toward structured output formats, fine-tuning ensures domain alignment, together producing results that neither approach alone achieves. This contributes to theoretical debates about how best to enforce structured outputs in narrative domains and supports the argument that lightweight, targeted adaptation can be a viable alternative to full retraining.

Third, applying this methodology to Chinese police briefings demonstrates the feasibility of bridging computational linguistics and criminology. Our work shows that narrative, domain-specific texts can be reliably converted into structured data, expanding the theoretical scope of IE into new applied domains.

### 4.4 Practical Implications

Beyond its theoretical contributions, this study offers several practical benefits. First, the proposed pipeline provides a cost-effective and scalable solution for researchers and institutions that lack the resources to deploy or fine-tune very large models. By leveraging LoRA adaptation, the approach reduces the technical and financial barriers to applying structured information extraction in real-world settings.

Second, our pipeline demonstrates that well-crafted task-specific prompts, combined with fine-tuned small-parameter models, can effectively handle complex structured-text extraction tasks. This approach enables deeper applications in domain-specific text structuring and annotation. For instance, in narrative policy framework information coding, researchers can develop optimal prompts based on disciplinary-specific annotation guidelines and theoretical requirements, then use minimal human-annotated data as training sets for model fine-tuning, thereby achieving efficient coding of large-scale textual data.

Third, the release of a high-quality annotated dataset directly addresses the scarcity of structured crime data in China. This resource enables criminologists, social scientists, and policy researchers to conduct empirical studies that were previously limited by a lack of accessible data.

Fourth, the pipeline supports practical applications in public safety management and policy evaluation. By transforming unstructured police briefings into structured, machine-readable data, the method facilitates tasks such as real-time incident monitoring, risk assessment, and longitudinal analysis of crime trends. These applications can directly inform decision-making by public security agencies and community organizations.

Finally, the study highlights a complementary role for smaller, fine-tuned models and larger, general-purpose models. While the former excel at producing structured, domain-specific outputs efficiently, the latter retain advantages in tasks that require subjective judgment. This suggests that hybrid deployment strategies could maximize both efficiency and interpretive capacity in applied contexts.

### 4.5 Distinction from Existing Work

Conventional approaches for information extraction in legal and social science domains have evolved from rule-based systems (Chiticariu et al., 2013) to specialized neural architectures. While domain-adapted models like Legal-BERT (Chalkidis et al., 2020) have significantly advanced performance on standard tasks such as named entity recognition, they still largely adhere to a one-model-per-task paradigm. This approach necessitates training and maintaining separate models for entity, relation, and event extraction, a process that is both computationally expensive and practically cumbersome for comprehensive, multi-faceted social science research.

The advent of Large Language Models (LLMs) has introduced a new paradigm, yet their off-the-shelf application often yields inconsistent results on complex, domain-specific narratives. Recent research has moved towards unifying extraction tasks through instruction-tuning, as exemplified by frameworks like InstructUIE (Gupta, 2023), which standardizes diverse extraction formats into a single text-to-structure generation task.

Our work builds on this trajectory but is new in two fundamental ways. First, we propose a unified pipeline specifically tailored to the deep semantic and contextual demands of socio-legal research. Unlike general-domain extraction, our tasks require sophisticated semantic interpretation—for instance, identifying a crime location from a descriptive narrative rather than a standardized field. Our integrated approach enables a single model to perform multiple structured extraction tasks (entities, relations, events), aligning with the holistic analytical needs of social scientists and mitigating the persistent challenge of limited annotated data in specialized domains. Second, we demonstrate the efficacy and practicality of our approach under significant resource constraints by leveraging parameter-efficient fine-tuning (PEFT). While state-of-the-art performance is often achieved through full fine-tuning, this method is computationally prohibitive for most researchers. By employing LoRA, we significantly reduce computational overhead while maintaining comparable performance. Implemented through the user-friendly LLaMA-Factory toolkit, our method lowers the technical barrier to entry, enabling social science researchers who may lack specialized coding expertise or access to high-performance computing resources to harness state-of-the-art LLMs for their analytical needs.

### 4.6 **Limitations and Future Research Directions**

We observed that the model underperforms on extraction tasks requiring professional knowledge or subjective judgment. For instance, in assessing severe social impacts, the fine-tuned Qwen2.5-7B-LoRA achieved only 68.18% accuracy, whereas the latest full-parameter models—DeepSeek-V3 and Qwen-Max—reached 88.86% and 91.6%, respectively. This indicates that newer base models align more closely with human value judgments, warranting further investigation. In the model evaluation section, only the accuracy of case-type classification was assessed, without an in-depth analysis of its predictive performance across case categories. We hope that future research will conduct more rigorous investigations into the discriminative capabilities of large language models across various case types, with the aim of mitigating model hallucinations and biases.

We further investigate fine-tuning LLMs for structuring and extracting information from domain-specific texts. While LLMs possess strong generalization and reasoning capabilities, they often lack domain expertise and exhibit variability in their outputs. To mitigate this, we constructed a dataset of police incident announcements and fine-tuned LoRA with prompt engineering. Experimental results demonstrate that fine-tuning Qwen-7B with a compact, high-quality dataset markedly improves extraction performance, even matching or exceeding that of full-parameter large models.

However, the model underperforms in several extraction tasks. It struggles to infer city-level locations from district names when broader contextual cues are absent. In pure-text extraction, responses exhibit high variability, resulting in low text-similarity scores—though semantic accuracy remains intact. Performance also remains suboptimal for case-type classification that requires domain expertise, likely due to insufficient or ambiguous professional descriptions in the source text. Moreover, on socially sensitive tasks that require subjective judgment, smaller, fine-tuned models still lag larger base models, raising questions about the viability of using LLMs to simulate human decision-making. The generalization capabilities of the fine-tuned compact model were not thoroughly validated, which may constrain its applicability beyond the specific dataset used (Zhang et al., 2021). Future research directions could integrate the fine-tuning technique with advanced methods such as reinforcement learning (Guo et al., 2025) or combine these techniques with AI agent technologies (Talebirad & Nadiri, 2023) to enhance model robustness and better support complex downstream tasks.

## 5. Conclusion

This study presents a novel approach to structured information extraction from police briefings, an important yet underutilized data source for criminology and public policy researchers. The core contribution is a domain-adapted extraction pipeline that integrates task-specific prompt engineering with LoRA fine-tuning of the Qwen2.5-7B model, providing a cost-effective and efficient means of transforming unstructured police briefings into structured, analyzable datasets. We rigorously benchmark this pipeline against baseline, instruction-tuned, and state-of-the-art models, showing clear improvements: 98.36% accuracy for mortality detection, 95.31% exact match rate for fatality counts, and 95.54% exact match rate for province-level location extraction. These results demonstrate the effectiveness and robustness of the proposed approach. Beyond its methodological contributions, this pipeline provides a scalable and accessible solution for criminological and social science research, enabling the systematic use of police briefings as structured data for further spatiotemporal analysis of crime or deviant behaviors.